\def\year{2021}\relax
\documentclass[letterpaper]{article} 
\usepackage{aaai21}  
\usepackage{times}  
\usepackage{helvet} 
\usepackage{courier}  
\usepackage[hyphens]{url}  
\usepackage{graphicx} 
\usepackage[table]{xcolor}

\usepackage{caption} 
\usepackage[square,numbers]{natbib}

\usepackage{graphicx}
\usepackage{xcolor}

\usepackage{amsmath,amssymb,amsfonts,amsthm} 

\definecolor{issuePJA_color}{rgb}{1.0,0.0,0.0}

\definecolor{commentPJA_color}{rgb}{1.0,0.0,0.8}

\definecolor{issueRL_color}{rgb}{1.0,0.8,0.0}

\definecolor{commentRL_color}{rgb}{0.6,0.0,0.8}

\definecolor{rev_color}{rgb}{0.6,0.0,0.0}

\newcommand{\mb}[1]{\mathbf{#1}}
\newcommand{\bsy}[1]{\boldsymbol{#1}}

\definecolor{atz_table1}{rgb}{0.85, 0.85, 0.85}
\definecolor{atz_table2}{rgb}{0.8, 0.8, 0.8}
\definecolor{atz_table3}{rgb}{0.75, 0.75, 0.75}

\title{Variational Autoencoders for Learning Nonlinear Dynamics of Physical Systems}

\author{
    Ryan Lopez,\textsuperscript{\rm 3} Paul J. Atzberger \textsuperscript{\rm 1,2,+}\thanks{Work supported by grants DOE Grant ASCR PHILMS DE-SC0019246 and NSF Grant DMS-1616353.} \\
}
\affiliations{
    \textsuperscript{\rm 1} Department of Mathematics, University of California Santa Barbara (UCSB). \\
    \textsuperscript{\rm 2}  Department of Mechanical Engineering, University of California Santa Barbara (UCSB). \\
    \textsuperscript{\rm 3} Department of Physics, University of California Santa Barbara (UCSB). \\
    \textsuperscript{\rm $+$}
    atzberg@gmail.com\\
    \url{http://atzberger.org/}\\
}

\begin{document}

\maketitle

\begin{abstract}
We develop data-driven methods for incorporating physical information for priors to learn parsimonious representations of nonlinear systems arising from parameterized PDEs and mechanics.  Our approach is based on Variational Autoencoders (VAEs) for learning nonlinear state space models from observations.  We develop ways to incorporate geometric and topological priors through general manifold latent space representations.  We investigate the performance of our methods for learning low dimensional representations for the nonlinear Burgers equation and constrained mechanical systems.  
\end{abstract}

\section{Introduction}
The general problem of learning dynamical models from a time series of observations has a long history spanning many fields
\cite{Nelles_Book_Nonlinear_Sys_Identification_2013,
Sjoeberg_Black_Box_Nonlinear_Dyn_1995,
Chiuso_System_Identification_Review_2019,
Hong_System_Identification_Review_2008} 
including in dynamical systems
\cite{Kutz_Brunton_book_ch_ROMs_2019,
Sjoeberg_Black_Box_Nonlinear_Dyn_1995,
Mallet_Coifman_Manifold_Learning_LVM_Dyn_Sys_2015,
Kutz_Lusch_Nonlinear_Embeddings_2018,
Mezic_Koopman_Review_2013,
Ohlberger_Redcuced_Basis_Review_2016,Hesthaven2016,
Crutchfield_Dynamics_Symbolic_1987,
DeVore_Reduced_Basis_Methods_Ch_2017},
control~\cite{Brunton_Kutz_SINDy_2016,
Nelles_Book_Nonlinear_Sys_Identification_2013,
Kutz_DNN_Time_Step_Constraints_2018,DMD_Schmid_2010}, 
statistics
\cite{Archer_Variational_State_Space_2015,
Jordan_Nonlinear_Dynamics_Theory_2020,
Ghahramani_EM_Nonlinear_Dynamics_1998},
and 
machine learning
\cite{Chiuso_System_Identification_Review_2019,
Hong_System_Identification_Review_2008,
Carlberg_Lee_Nonlinear_Dynamics_AE_2020,
Karniadakis_Raissi_Hidden_Phys_PDEs_2018,
Bertozzi_CDMD_2019,
Perdikaris_Phys_Informed_Gen_Models_2018}.  Referred to as system identification in control and engineering, many approaches have been developed starting with linear dynamical systems (LDS).
These includes the Kalman Filter and extensions~\cite{Kalman1960,
DelMoral1997,
Godsill2019,
VanDerMerwe_Kalman_Unscented_2000,
Wan_Kalman_Nonlinear_2000}, Principle Orthogonal Decomposition (POD)~\cite{POD_Intro_Chatterjee_2000,Mendez2018}, and more recently Dynamic Mode Decomposition (DMD)
\cite{DMD_Schmid_2010,
DMD_Kutz_Brunton_Book_2016,
DMD_Theory_and_App_Kutz_2014}
and Koopman Operator approaches
\cite{Mezic_Koopman_Review_2013,Das2019,
Putinar_Mezic_cont_spectrum_Koopman_2020}.  
These successful and widely-used approaches rely on assumptions on the model structure, most commonly, that a time-invariant LDS provides a good local approximation or that noise is Gaussian. 

There also has been research on 
more general nonlinear system identification 
\cite{Archer_Variational_State_Space_2015,
Ljung_Nonlinear_Sys_Iden_Review_2019,
Chiuso_System_Identification_Review_2019,
Hong_System_Identification_Review_2008,
Schoen_ML_State_Space_Est_2011,
Kutz_Lusch_Nonlinear_Embeddings_2018,
Jordan_Nonlinear_Dynamics_Theory_2020,
Nelles_Book_Nonlinear_Sys_Identification_2013}.  
Nonlinear systems pose many open challenges and fewer unified approaches given the rich behaviors of nonlinear dynamics.  For classes of systems and specific application domains, methods have been developed which make different levels of assumptions about the underlying structure of the dynamics.  Methods for learning nonlinear dynamics include the NARAX and NOE approaches with function approximators based on neural networks and other models classes
\cite{Nelles_Book_Nonlinear_Sys_Identification_2013,
Sjoeberg_Black_Box_Nonlinear_Dyn_1995}, 
sparse symbolic dictionary methods that are linear-in-parameters such as SINDy
\cite{Brunton_Kutz_SINDy_2016,
Schmidt2009,
Sjoeberg_Black_Box_Nonlinear_Dyn_1995}, 
and dynamic Bayesian networks (DBNs), such as
Hidden Markov Chains (HMMs) and Hidden-Physics Models
~\cite{Karniadakis_Raissi_Hidden_Phys_PDEs_2018,
Pawar2020,Saul2020,Baum_HMM_1966,Krishnan_GMS_2017,
Ghahramani_EM_Nonlinear_Dynamics_1998}.  

A central challenge in learning non-linear dynamics is to obtain representations not only capable of reproducing similar outputs as observed directly in the training dataset but to infer structures that can provide stable more long-term extrapolation capabilities over multiple future steps and input states.  
In this work, we develop learning methods aiming to obtain robust non-linear models by providing ways to incorporate more structure and information about the underlying system related to smoothness, periodicity, topology, and other constraints.  We focus particularly on developing Probabilistic Autoencoders (PAE) that incorporate noise-based regularization and priors to learn lower dimensional representations from observations.  This provides the basis of nonlinear state space models for prediction.  We develop methods for incorporating into such representations geometric and topological information about the system.  This facilitates capturing qualitative features of the dynamics to enhance robustness and to aid in interpretability of results.  We demonstrate and perform investigations of our methods to obtain models for reductions of parameterized PDEs and for constrained mechanical systems.

\section*{Learning Nonlinear Dynamics with \\ Variational Autoencoders (VAEs)}
\label{sec_VAR}

We develop data-driven approaches based on a Variational Autoencoder (VAE) framework~\cite{KingmaWellingVAE2014}.  We learn from observation data a set of lower dimensional representations that are used to make predictions for the dynamics.  In practice, data can include experimental measurements, large-scale computational simulations, or solutions of complicated dynamical systems for which we seek reduced models.  Reductions aid in gaining insights for a class of inputs or physical regimes into the underlying mechanisms generating the observed behaviors.  Reduced descriptions are also helpful in many optimization problems in design and in development of controllers
\cite{Nelles_Book_Nonlinear_Sys_Identification_2013}.  

Standard autoencoders can result in encodings that yield
unstructured scattered disconnected coding points for system features $\mb{z}$.  VAEs provide probabilistic encoders and decoders where noise provides regularizations that promote more connected encodings, smoother dependence on inputs, and more disentangled feature components~\cite{KingmaWellingVAE2014}.  As we shall discuss, we also introduce other regularizations into our methods to help aid in interpretation of the learned latent representations.  

\begin{figure}[ht]
\centerline{\includegraphics[width=0.99\columnwidth]{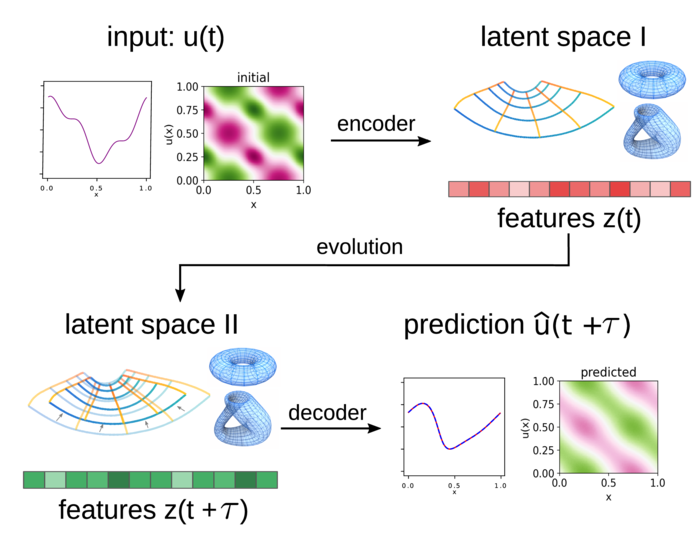}}
\caption{\textbf{Learning Nonlinear Dynamics.}  Data-driven methods are developed for learning robust models to predict from  $u(x,t)$ the non-linear evolution to $u(x,t + \tau)$ for PDEs and other dynamical systems.  Probabilistic Autoencoders (PAEs) are utilized to learn representations $z$ of $u(x,t)$ in low dimensional latent spaces with prescribed geometric and topological properties.  The model makes predictions using learnable maps that (i) encode an input $u(x,t) \in \mathcal{U}$ as $z(t)$ in latent space \textit{(top)}, (ii) evolve the representation $z(t) \rightarrow z(t + \tau)$ \textit{(top-right)}, (iii) decode the representation $z(t + \tau)$ to predict $\hat{u}(x,t + \tau)$ \textit{(bottom-right)}.
}
\label{fig:vae_learning_dynamics}
\end{figure}

\begin{figure}[ht]
\centerline{\includegraphics[width=0.99\columnwidth]{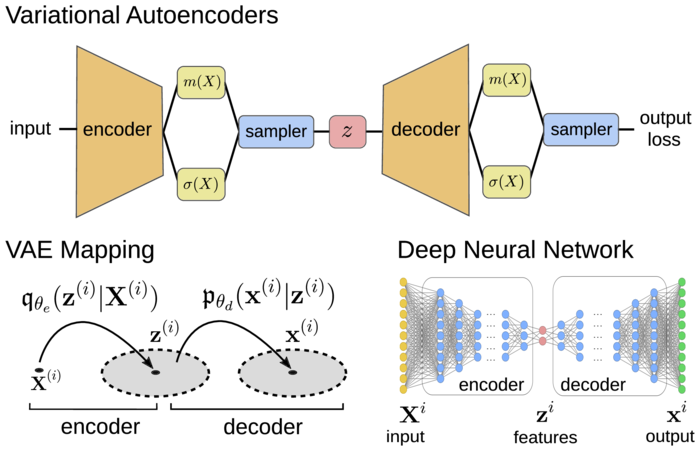}}
\caption{\textbf{Variational Autoencoder (VAE).}  VAEs~\cite{KingmaWellingVAE2014} are used to learn representations of the nonlinear dynamics.  Deep Neural Networks (DNNs) are trained (i) to serve as feature extractors to represent functions $u(x,t)$ and their evolution in a low dimensional latent space as $z(t)$  (encoder $\sim \mathfrak{q}_{\theta_e}$), and (ii) to serve as approximators that can construct predictions $u(x,t + \tau)$ using features $z(t+ \tau)$ (decoder $\sim \mathfrak{p}_{\theta_d}$).  
}
\label{VAE_schematic}
\end{figure}

We learn VAE predictors using a Maximum Likelihood Estimation (MLE) approach for the Log Likelihood (LL) $\mathcal{L}_{LL} = \log(p_\theta(\mb{X},\mb{x}))$.  For dynamics of $u(s)$, let $\mb{X} = u(t)$ and $\mb{x} = u(t + \tau)$.  We base $p_\theta$ on the autoencoder framework in Figure~\ref{fig:vae_learning_dynamics} and~\ref{VAE_schematic}.  We use variational inference to approximate the LL by the Evidence Lower Bound (ELBO)~\cite{Blei_VI_Review_2017} to train a model with parameters $\theta$ using encoders and decoders based on minimizing the loss function
\begin{eqnarray}
\nonumber
\theta^* &=& \arg\min_{\theta_e,\theta_d} -\mathcal{L}^B(\theta_e,\theta_d,\theta_\ell;\mb{X}^{(i)},\mb{x}^{(i)}), \\
\label{equ:vae_loss}
\mathcal{L}^B &=&  \mathcal{L}_{RE} + \mathcal{L}_{KL} + \mathcal{L}_{RR}, \\
\nonumber
\mathcal{L}_{RE} &=&
E_{\mathfrak{q}_{\theta_e}(\mb{z}|\mb{X}^{(i)})}\left\lbrack \log \mathfrak{p}_{\theta_d}(\mb{x}^{(i)} | \mb{z}') \right\rbrack  \\
\nonumber
\mathcal{L}_{KL} &=&
-\beta\mathcal{D}_{KL}\left(\mathfrak{q}_{\theta_e}(\mb{z}|\mb{X}^{(i)}) \, \| \, 
\tilde{\mathfrak{p}}_{\theta_d}(\mb{z})\right) \\
\nonumber
\mathcal{L}_{RR} &=&
\gamma 
E_{\mathfrak{q}_{\theta_e}(\mb{z}'|\mb{x}^{(i)})}\left\lbrack \log \mathfrak{p}_{\theta_d}(\mb{x}^{(i)} | \mb{z}') \right\rbrack.
\end{eqnarray}
The $\mathfrak{q}_{\theta_e}$ denotes the encoding probability distribution and $\mathfrak{p}_{\theta_d}$ the decoding probability distribution.  The loss $\ell = -\mathcal{L}^B$ provides a regularized form of MLE. 

The terms $\mathcal{L}_{RE}$ and $\mathcal{L}_{KL}$ arise from the ELBO variational bound $\mathcal{L}_{LL} \geq \mathcal{L}_{RE} + \mathcal{L}_{KL}$ when $\beta = 1$,~\cite{Blei_VI_Review_2017}. This provides a way to estimate the log likelihood that the encoder-decoder reproduce the observed data sample pairs $(\mb{X}^{(i)},\mb{x}^{(i)})$ using the codes $\mb{z'}$ and $\mb{z}$.  Here, we include a latent-space mapping $\mb{z}' = f_{\theta_\ell}(\mb{z})$ parameterized by $\theta_\ell$, which we can use to characterize the evolution of the system or further processing of features. The $\mb{X}^{(i)}$ is the input and $\mb{x}^{(i)}$ is the output prediction.  For the case of dynamical systems, we take $\mb{X}^{(i)} \sim u^i(t)$ a sample of the initial state function $u^i(t)$ and the output $\mb{x}^{(i)} \sim u^i(t + \tau)$ the predicted state function $u^i(t + \tau)$.  We discuss the specific distributions used in more detail below.

The $\mathcal{L}_{KL}$ term involves the Kullback-Leibler Divergence
\cite{Kullback1951,Book_Cover_Thomas_2006} acting similar to a Bayesian prior on latent space to regularize the encoder conditional probability distribution so that for each sample this distribution is similar to $p_{\theta_d}$.  We take $p_{\theta_d} = \eta(0,\sigma_0^2)$ a multi-variate Gaussian with independent components.  This serves (i) to disentangle the features from each other to promote independence, (ii) provide a reference scale and localization for the encodings $\mb{z}$, and (iii) promote parsimonious codes utilizing smaller dimensions than $d$ when possible.

The $\mathcal{L}_{RR}$ term gives a regularization that promotes retaining information in $z$ so the encoder-decoder pair can reconstruct functions.  As we shall discuss, this also promotes organization of the latent space for consistency over multi-step predictions and aids in model interpretability.

We use for the specific encoder probability distributions conditional Gaussians
$\mb{z} \sim \mathfrak{q}_{\theta_e}(\mb{z}|\mb{x}^{(i)}) = \mathfrak{a}(\mb{X}^{(i)},\mb{x}^{(i)}) + \eta(0,\sigma_e^2)$ where 
$\eta$ is a Gaussian with variance $\sigma_e^2$, (i.e. $\mathbb{E}^{\mb{X}^i}[z] = \mathfrak{a}$, 
$\mbox{Var}^{\mb{X}^i}[z] = \sigma_e^2$). One can think of the learned mean function  $\mathfrak{a}$ in the VAE as corresponding to a typical 
encoder 
$\mathfrak{a}(\mb{X}^{(i)},\mb{x}^{(i)};\theta_e) = \mathfrak{a}(\mb{X}^{(i)};\theta_e) = \mb{z}^{(i)}$ 
and the variance function $\sigma_e^2 = \sigma_e^2(\theta_e)$ as providing control of a noise source to further regularize the encoding.  Among other properties, this promotes connectedness of the ensemble of latent space codes.  For the VAE decoder distribution, we take $\mb{x} \sim \mathfrak{p}_{\theta_d}(\mb{x}|\mb{z}^{(i)}) = \mathfrak{b}(\mb{z}^{(i)}) + \eta(0,\sigma_d^2)$.  The learned mean function $\mathfrak{b}(\mb{z}^{(i)};\theta_e)$ 
corresponds to a typical decoder and the variance function $\sigma_e^2 = \sigma_e^2(\theta_d)$ controls the source of regularizing noise.  

The terms to be learned in the VAE framework are $(\mathfrak{a},\sigma_e,f_{\theta_\ell},\mathfrak{b},\sigma_d)$ which are parameterized by $\theta = (\theta_e,\theta_d,\theta_\ell)$.  In practice, it is useful to treat variances $\sigma_{(\cdot)}$ initially as hyper-parameters.  
We learn predictors for the dynamics by training over samples of evolution pairs $\{(u_n^i,u_{n+1}^i)\}_{i=1}^m$, where $i$ denotes the sample index and 
$u_n^i = u^i(t_n)$ with $t_n = t_0 + n\tau$ for a time-scale $\tau$. 

To make predictions, the learned models use the following stages: (i) extract from $u(t)$ the features $z(t)$, (ii) evolve $z(t) \rightarrow z(t + \tau)$, (iii) predict 
using $z(t + \tau)$ the $\hat{u}(t + \tau)$, summarized in Figure~\ref{fig:vae_learning_dynamics}.  By composition of the latent evolution map the model makes multi-step predictions of the dynamics.

\section{Learning with Manifold Latent Spaces \\ Roles of Non-Euclidean Geometry and Topology}
For many systems, parsimonious representations can be obtained by working with non-euclidean manifold latent spaces, such as a torus for doubly periodic systems or even non-orientable manifolds, such as a klein bottle as arises in imaging and perception studies~\cite{Carlsson2008}.  For this purpose, we learn encoders $\mathcal{E}$ over a family of mappings to a prescribed manifold $\mathcal{M}$ of the form
$$\mb{z} = \mathcal{E}_\phi(\mb{x}) = \Lambda(\tilde{\mathcal{E}}_{\phi}(\mb{x})) = \Lambda(\mb{w}), \; \; \; \mb{w} = \tilde{\mathcal{E}}_{\phi}(\mb{x}).$$  We take the map $\tilde{\mathcal{E}}_{\phi}(\mb{x}): \mb{x} \rightarrow \mb{w}$, where we represent a smooth closed manifold $\mathcal{M}$ of dimension $m$ in $\mathbb{R}^{2m}$, as supported by the Whitney Embedding Theorem
\cite{Whitney1944}. The $\Lambda$ maps (projects) points $\mb{w} \in \mathbb{R}^{2m}$ to the manifold representation $\mb{z} \in \mathcal{M} \subset \mathbb{R}^{2m}$.  In practice, we accomplish this two ways: (i) we provide an analytic mapping $\Lambda$ to $\mathcal{M}$, (ii) we provide a high resolution point-cloud representation of the target manifold along with local gradients and use for $\Lambda$ a quantized mapping to the nearest point on $\mathcal{M}$.  We provide more details in Appendix A.  

This allows us to learn VAEs with latent spaces for $\mb{z}$ with general specified topologies and controllable geometric structures.  The topologies of sphere, torus, klein bottle are intrinsically different than $\mathbb{R}^n$.  This allows for new types of priors such as uniform on compact manifolds or distributions with more symmetry.
As we shall discuss, additional latent space structure also helps in learning more robust representations less sensitive to noise since we can unburden the encoder and decoder from having to learn the embedding geometry and avoid the potential for them making erroneous use of  extra latent space dimensions.  We also have statistical gains since the decoder now only needs to learn a mapping from the manifold $\mathcal{M}$ for reconstructions of $\mb{x}$.  These more parsimonious representations also aid identifiability and interpretability of models.

\section{Related Work}
Many variants of autoencoders have been developed for making predictions of sequential data, including those based on
Recurrent Neural Networks (RNNs) with LSTMs and GRUs~\cite{Schmidhuber_LSTM_1997,
Goodfellow2016,Cho_Bengio_GRU_2014}.  While RNNs provide a rich approximation class for sequential data, they pose for dynamical systems challenges for interpretability and for training to obtain predictions stable over many steps with robustness against noise in the training dataset.  Autoencoders have also been combined with symbolic dictionary learning 
for latent dynamics in \cite{Kutz_Discovery_Coordinates_2019} providing some advantages for interpretability and robustness, but require specification in advance of a sufficiently expressive dictionary.  Neural networks incorporating physical information have also been developed that impose stability conditions during training
\cite{Carlberg_RNN_Dynamics_2020,
Carlberg_Lee_Nonlinear_Dynamics_AE_2020,
Erichson_AE_Lyapunov_Stable_Flow_2019}.  The work of \cite{Bengio_Recurrent_LVM_2015} 
investigates combining RNNs with VAEs to obtain more robust models for sequential data and considered tasks related to processing speech and handwriting.  

In our work we learn dynamical models making use of VAEs to obtain probabilistic encoders and decoders between euclidean and non-euclidean latent spaces to provide additional regularizations to help promote parsimoniousness, disentanglement of features, robustness, and interpretability.  Prior VAE methods used for dynamical systems include
\cite{Hernandez_VAE_Dynamics_2018,
Pearce_Latent_Dynamics_GP_2020,
Girin_Dynamical_VAE_Review_2020,
Chen_VAE_Dynamic_Motion_Primitives_2016,
Pearce_Latent_Dynamics_GP_2020,
Roeder_VAE_Dynamics_Hierarchical_2019}.
These works use primarily euclidean latent spaces and consider applications including human motion capture and ODE systems.
Approaches for incorporating topological information into latent variable representations include the early works by Kohonen on Self-Organizing Maps (SOMs)
\cite{Kohonen_Original_Paper_Self_Organizing_Maps_1982}
and Bishop on Generative Topographical Maps (GTMs)
based on density networks providing a generative approach 
\cite{Bishop_Early_Generative_Topographical_Map_1996}.  More recently, VAE methods using non-euclidean latent spaces include 
\cite{Jensen_Manifold_Latent_Model_2020,
Kalatzis_VAE_B_Motion_Priors_2020,
Falorsi_Homeomorphic_VAE_2018,
Chen_VAE_Learn_Flat_Manifolds_2020,
Kipf_VAE_Hyperspherical_2018,
Arvanitidis_Noneuclidean_Latent_Space_2018}.
These incorporate the role of geometry by augmenting the prior distribution $\tilde{\mathfrak{p}}_{\theta_d}(z)$ on latent space to bias toward a manifold.  In the recent work
\cite{Rey_Diffusion_VAE_2020}, an explicit projection procedure is introduced, but in the special case of a few manifolds having an analytic projection map.

In our work we develop further methods for more general latent space representations, including non-orientable manifolds, and applications to parameterized PDEs and constrained mechanical systems.  We introduce more general methods for non-euclidean latent spaces in terms of point-cloud representations of the manifold along with local gradient information that can be utilized within general back-propogation frameworks, see Appendix A.  This also allows for the case of manifolds that are non-orientable and having complex shapes.  Our methods provide flexible ways to design and control both the topology and the geometry of the latent space by merging or subtracting shapes or stretching and contracting regions.  We also consider additional types of regularizations for learning dynamical models facilitating multi-step predictions and more interpretable state space models.  In our work, we also consider reduced models for non-linear PDEs, such as Burgers Equations, and learning representations for more general constrained mechanical systems.  We also investigate the role of non-linearities making comparisons with other data-driven models.

\section*{Results}
\label{sec_Results}

\subsection*{Burgers' Equation of Fluid Mechanics: Learning Nonlinear PDE Dynamics}

We consider the nonlinear viscous Burgers' equation
\begin{eqnarray}
u_t = -uu_x + \nu u_{xx},
\label{eqn:burgers}
\end{eqnarray}
where $\nu$ is the viscosity~\cite{Bateman1915,Hopf1950}.  We consider periodic boundary conditions on $\Omega = [0,1]$.  Burgers equation is motivated as a mechanistic model for the fluid mechanics of advective transport and shocks, and serves as a widely used benchmark for analysis and computational methods.

The nonlinear Cole-Hopf Transform $\mathcal{CH}$ can be used to relate Burgers equation to the linear Diffusion equation $\phi_t = \nu \phi_{xx}$~\cite{Hopf1950}.  This provides a representation of the solution $u$ 
\begin{eqnarray}
\nonumber
\phi(x,t) & = & \mathcal{CH}[u] = \exp \left( -\frac{1}{2\nu}\int_0^x u(x',t)dx'\right) \\
u(x,t) & = & \mathcal{CH}^{-1}[\phi] = -2\nu \frac{\partial}{\partial x} \ln \phi(x,t).
\end{eqnarray}
This can be represented by the Fourier expansion
\begin{eqnarray}
\nonumber
&& \phi(x,t) = \sum_{k=-\infty}^{\infty} \hat{\phi}_k(0) \exp(-4\pi^2 k^2 \nu t) \cdot \exp(i2\pi kx).
\end{eqnarray}
The $\hat{\phi}_k(0) = \mathcal{F}_k[\phi(x,0)]$ and $\phi(x,t) = \mathcal{F}^{-1}[\{\hat{\phi}_k(0) \exp(-4\pi^2 k^2 \nu t)\}]$ with $\mathcal{F}$ the Fourier transform. This provides an analytic representation of the solution of the viscous Burgers equation $u(x,t) = \mathcal{CH}^{-1}[\phi(x,t)]$ where $\hat{\phi}(0) = \mathcal{F}[\mathcal{CH}[u(x,0)]]$.  In general, for nonlinear PDEs with initial conditions within a class of functions $\mathcal{U}$, we aim to learn models that provide predictions $u(t + \tau) = \mathcal{S}_\tau u(t)$ approximating the evolution operator $\mathcal{S}_\tau$ over time-scale $\tau$.  For the Burgers equation, the $\mathcal{CH}$ provides an analytic way to obtain a reduced order model by truncating the Fourier expansion to $|k| \leq n_f/2$.  This provides for the Burgers equation a benchmark model against which to compare our learned models. For general PDEs comparable analytic representations are not usually available, motivating development of data-driven approaches.

\begin{figure}[ht]
\centerline{\includegraphics[width=0.99\columnwidth]{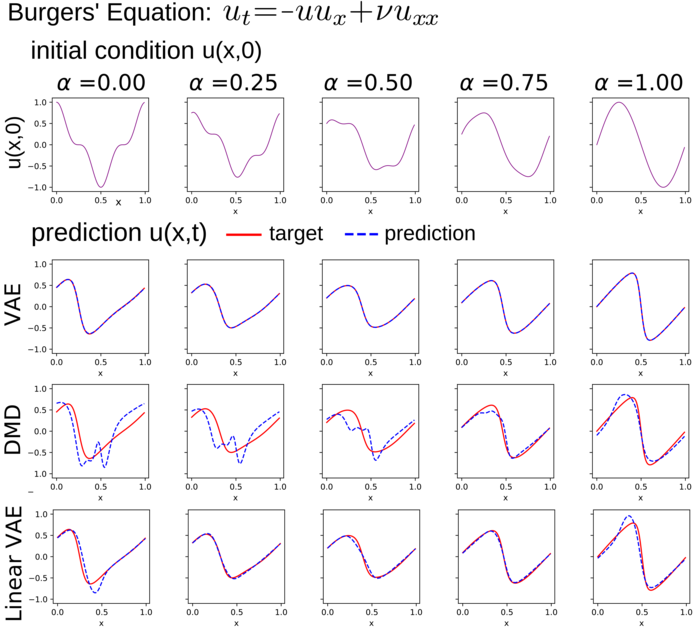}}  
\caption{\textbf{Burgers' Equation: Prediction of Dynamics.}  We consider responses for $\mathcal{U}_1 = \{ u\, | \, u(x,t;\alpha) = \alpha \sin(2\pi x) + (1-\alpha) \cos^3(2 \pi x)\}$.  Predictions are made for the evolution $u$ over the time-scale $\tau$ satisfying equation~\ref{eqn:burgers} with initial conditions in $\mathcal{U}_1$.  We find our nonlinear VAE methods are able to learn with $2$ latent dimensions the dynamics with errors $< 1\%$. Methods such as DMD~\cite{DMD_Schmid_2010,DMD_Theory_and_App_Kutz_2014} with $3$ modes which are only able to use a single linear space to approximate the initial conditions and prediction encounter challenges in approximating the nonlinear evolution.  We find our linear VAE method with $2$ modes provides some improvements, by allowing for using different linear spaces for representing the input and output functions, but at the cost of additional computations.  Results are summarized in Table~\ref{table:Burgers_compare_VAE_DMD_etc}.}
\label{fig:Burgers_compare_VAE_DMD_etc}
\end{figure}

We develop VAE methods for learning reduced order models for the responses of nonlinear Burgers Equation when the initial conditions are from a collection of functions $\mathcal{U}$.  We learn VAE models that extract from 
$u(x,t)$ latent variables $z(t)$ to predict $u(x,t + \tau)$.  Given the non-uniqueness of representations and to promote interpretability of the model, we introduce the inductive bias that the evolution dynamics in latent space for $z$ is linear of the form $\dot{z} = -\lambda_0 z$, giving exponential decay rate $\lambda_0$.  For discrete times, we take $z_{n+1} = f_{\theta_\ell}(z_n) = \exp(-\lambda_0 \tau) \cdot z_n$, where $\theta_\ell = (\lambda_0)$.  We still consider general nonlinear mappings for the encoders and decoders which are represented by deep neural networks.  We train the model on the pairs $(u(x,t),u(x,t + \tau))$ by drawing $m$ samples of $u^{i}(x,t_i) \in \mathcal{S}_{t_i} \mathcal{U}$ which generates the evolved state under Burgers equation $u^i(x,t_i + \tau)$ over time-scale $\tau$.  
We perform VAE studies with parameters $\nu = 2 \times 10^{-2}$, $\tau = 2.5 \times 10^{-1}$ with VAE Deep Neural Networks (DNNs) with layer sizes (in)-400-400-(out), ReLU activations, and $\gamma = 0.5$, $\beta = 1$, and initial standard deviations $\sigma_d = \sigma_e = 4 \times 10^{-3}$.
We show results of our VAE model predictions in Figure~\ref{fig:Burgers_compare_VAE_DMD_etc} and Table~\ref{table:Burgers_compare_VAE_DMD_etc}.

\begin{figure}[ht]
\centerline{\includegraphics[width=0.99\columnwidth]{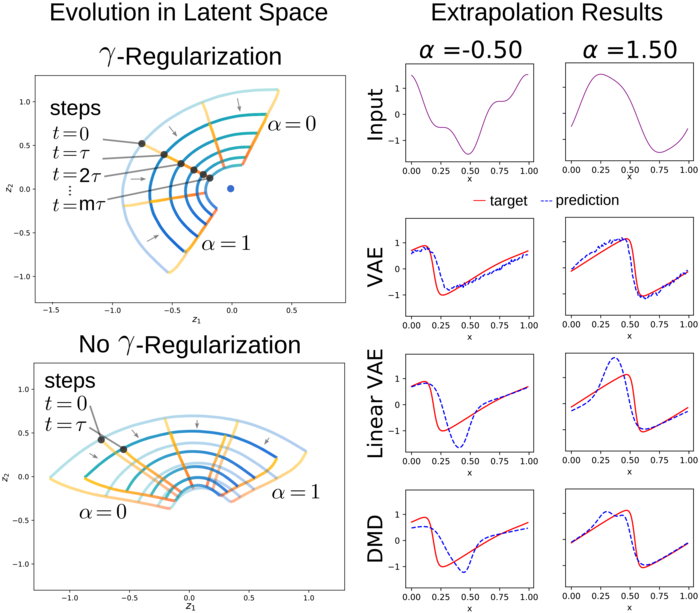}} 
\caption{\textbf{Burgers' Equation: Latent Space Representations and Extrapolation Predictions.}  We show the latent space representation $z$ of the dynamics for the input functions $u(\cdot,t;\alpha) \in \mathcal{U}_1$.  VAE organizes for $u$ the learned representations $z(\alpha,t)$  in parameter $\alpha$ \textit{(blue-green)} into circular arcs that are concentric in the time parameter $t$, \textit{(yellow-orange)} \textit{(left)}.  The reconstruction regularization with $\gamma$ aligns subsequent time-steps of the dynamics in latent space facilitating multi-step predictions.  The learned VAE model exhibits a level of extrapolation to predict dynamics even for some inputs $u \not \in \mathcal{U}_1$ beyond the training dataset \textit{(right)}.
}
\label{fig:burgers_latent_space}	
\label{fig:Burgers_compare_VAE_DMD_etc_outside_training}
\end{figure}

We show the importance of the non-linear approximation properties of our VAE methods in capturing system behaviors by making comparisons with Dynamic Mode Decomposition (DMD)~\cite{DMD_Schmid_2010,DMD_Theory_and_App_Kutz_2014}, Principle Orthogonal Decomposition (POD)~\cite{POD_Intro_Chatterjee_2000}, and a linear variant of our VAE approach.  Recent CNN-AEs have also studied related advantages of non-linear approximations~\cite{Carlberg_Lee_Nonlinear_Dynamics_AE_2020}.  Some distinctions in our work is the use of VAEs to further regularize AEs and using topological latent spaces to facilitate further capturing of structure.  The DMD and POD are widely used and successful approaches that aim to find an optimal linear space on which to project the dynamics and learn a linear evolution law for system behaviors.  DMD and POD have been successful in obtaining models for many applications, including steady-state fluid mechanics and transport problems
\cite{DMD_Theory_and_App_Kutz_2014,
DMD_Schmid_2010}.  However, given their inherent linear approximations they can encounter well-known challenges related to translational and rotational invariances, as arise in advective phenomena and other settings~\cite{Kutz_Brunton_book_ch_ROMs_2019}.    Our comparison studies can be found in Table~\ref{table:Burgers_compare_VAE_DMD_etc}.

\begin{table}[ht]
\begin{center}
{
\fontsize{8.3}{10}\selectfont
\selectfont
\begin{tabular}{l|c|cccc}
\hline 
\rowcolor{atz_table1} 
\textbf{Method} & \textbf{Dim} & \textbf{0.25s} & \textbf{0.50s} & \textbf{0.75s} & \textbf{1.00s} \\
\cline{1-6}
VAE Nonlinear  & 2 & \textbf{4.44e-3} & \textbf{5.54e-3} & \textbf{6.30e-3} & \textbf{7.26e-3} \\
VAE Linear  & 2 & 9.79e-2 & 1.21e-1 & 1.17e-1 & 1.23e-1\\
DMD & 3 & 2.21e-1 & 1.79e-1 & 1.56e-1 & 1.49e-1 \\
POD & 3 & 3.24e-1 & 4.28e-1 & 4.87e-1 & 5.41e-1\\
\cline{1-6}
Cole-Hopf-2 & 2 &5.18e-1 & 4.17e-1 &3.40e-1 & 1.33e-1\\
Cole-Hopf-4 & 4 & 5.78e-1 & 6.33e-2 & 9.14e-3 & 1.58e-3\\
Cole-Hopf-6 & 6 &1.48e-1 & 2.55e-3 & 9.25e-5 & 7.47e-6\\
\hline
\end{tabular} \\
}
\vspace{0.2cm}
{
\fontsize{7.6}{10}\selectfont
\begin{tabular}{c|ccccc}
\hline
\rowcolor{atz_table1} 
$\bsy{\gamma}$ & \textbf{0.00s} & \textbf{0.25s} & \textbf{0.50s} & \textbf{0.75s} & \textbf{1.00s} \\
\cline{1-6}
0.00 & 1.600e-01 & 6.906e-03 & 1.715e-01 & 3.566e-01 & 5.551e-01\\
0.50 & 1.383e-02 & 1.209e-02 & 1.013e-02 & 9.756e-03 & 1.070e-02\\
2.00 & 1.337e-02 & 1.303e-02 & 9.202e-03 & 8.878e-03 & 1.118e-02\\
\hline
\end{tabular}
} \\
\vspace{0.2cm}
{
\fontsize{7.6}{10}\selectfont
\begin{tabular}{c|ccccc}
\hline
\rowcolor{atz_table1}  
$\bsy{\beta}$ & \textbf{0.00s} & \textbf{0.25s} & \textbf{0.50s} & \textbf{0.75s} & \textbf{1.00s} \\
\cline{1-6}
0.00 & 1.292e-02 & 1.173e-02 & 1.073e-02 & 1.062e-02 & 1.114e-02\\
0.50 & 1.190e-02 & 1.126e-02 & 1.072e-02 & 1.153e-02 & 1.274e-02\\
1.00 & 1.289e-02 & 1.193e-02 & 7.903e-03 & 7.883e-03 & 9.705e-03\\
4.00 & 1.836e-02 & 1.677e-02 & 8.987e-03 & 8.395e-03 & 8.894e-03\\
\hline
\end{tabular}
}
\end{center}
\caption{\textbf{Burgers' Equation: Prediction Accuracy.} 
The reconstruction $L^1$-relative errors in predicting $u(x,t)$ for our VAE methods, Dynamic Model Decomposition (DMD), and Principle Orthogonal Decomposition (POD), and reduction by Cole-Hopf (CH), over multiple-steps and number of latent dimensions (Dim)
\textit{(top)}.  Results when varying the strength of the reconstruction regularization $\gamma$ and prior $\beta$ \textit{(bottom)}.
}
\label{table:Burgers_compare_VAE_DMD_etc}
\end{table}

We also considered how our VAE methods performed when adjusting the parameters $\beta$ for the strength of the prior $\tilde{\mathfrak{p}}$ as in $\beta$-VAEs~\cite{Higgins2017BetaVAE} and 
$\gamma$ for the strength of the reconstruction regularization.  The reconstruction regularization has a significant influence 
on how the VAE organizes representations in latent space and the accuracy of predictions of the dynamics, especially over multiple steps, see Figure~\ref{fig:Burgers_compare_VAE_DMD_etc_outside_training} and Table~\ref{table:Burgers_compare_VAE_DMD_etc}.     The regularization serves to align representations consistently in latent space facilitating multi-step compositions.  We also found our VAE learned representions capable of some level of extrapolation beyond the training dataset.
When varying $\beta$, we found that larger values improved the multiple step accuracy whereas small values improved the single step accuracy, see Table~\ref{table:Burgers_compare_VAE_DMD_etc}.

\subsection*{Constrained Mechanics: Learning with Non-Euclidean Latent Spaces}

\begin{figure}[ht]
\centerline{\includegraphics[width=0.99\columnwidth]{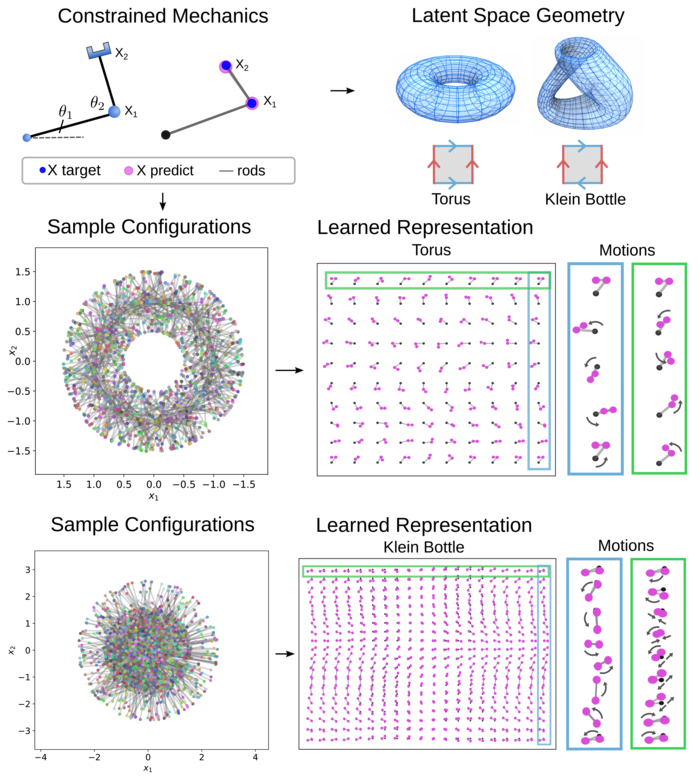}}
\caption{\textbf{VAE Representations of Motions using Manifold Latent Spaces.}  We learn from observations representations for constrained mechanical systems using general non-euclidean manifolds latent spaces $\mathcal{M}$.  The arm mechanism has configurations $\mb{x} = (\mb{x}_1,\mb{x}_2) \in \mathbb{R}^4$.  For rigid segments, the motions are constrained to be on a manifold (torus) $\mathcal{M} \subset \mathbb{R}^4$.  For extendable segments, we can also consider more exotic constraints, such as requiring $\mb{x}_1,\mb{x}_2$ to be on a klein bottle in $\mathbb{R}^4$ \textit{(top)}.  Results of our VAE methods for learned representations for motions under these constraints are shown.  VAE learns the segment length constraint and two nearly decoupled coordinates for the torus dataset that mimic the roles of angles.  VAE learns for the klein bottle dataset two segment motions to generate configurations \textit{(middle and bottom)}.
}
\label{fig:manifold_lvms}
\end{figure}

\begin{table}[ht]
\begin{center}
\fontsize{7.3}{10}
\selectfont
\begin{tabular}{l|cccc}
\hline
\rowcolor{atz_table3} 
\multicolumn{1}{l}{\textbf{Torus}} & 
\multicolumn{4}{|l}{epoch} \\
\hline
\rowcolor{atz_table1} 
method & 1000 & 2000 & 3000 & final\\
\hline
VAE 2-Manifold & \textbf{6.6087e-02} & \textbf{6.6564e-02} & \textbf{6.6465e-02} &  
\textbf{6.6015e-02}
\\
VAE $\mathbb{R}^2$ & 1.6540e-01 & 1.2931e-01 & 9.9903e-02 & 8.0648e-02 \\
VAE $\mathbb{R}^4$ & 8.0006e-02 & 7.6302e-02 & 7.5875e-02 & 7.5626e-02\\
VAE $\mathbb{R}^{10}$ & 8.3411e-02 & 8.4569e-02 & 8.4673e-02 & 8.4143e-02\\
\hline
\rowcolor{atz_table1} 
with noise $\sigma$ & 0.01 & 0.05 & 0.1 & 0.5 \\
\hline
VAE 2-Manifold & \textbf{6.7099e-02} & \textbf{8.0608e-02} & \textbf{1.1198e-01} & \textbf{4.1988e-01} \\
VAE $\mathbb{R}^2$ & 8.5879e-02 & 9.7220e-02 & 1.2867e-01 & 4.5063e-01 \\
VAE $\mathbb{R}^4$ & 7.6347e-02 & 9.0536e-02 & 1.2649e-01 & 4.9187e-01 \\
VAE $\mathbb{R}^{10}$ & 8.4780e-02 & 1.0094e-01 & 1.3946e-01 & 5.2050e-01 \\
\hline
\rowcolor{atz_table3}
\multicolumn{1}{l}{\textbf{Klein Bottle}} & 
\multicolumn{4}{|l}{epoch} \\
\hline
\rowcolor{atz_table1} 
method & 1000 & 2000 & 3000 & final\\
\cline{1-5}
VAE 2-Manifold & \textbf{5.7734e-02} & \textbf{5.7559e-02} & \textbf{5.7469e-02} & \textbf{5.7435e-02} \\
VAE $\mathbb{R}^2$ & 1.1802e-01 & 9.0728e-02 & 8.0578e-02 & 7.1026e-02 \\
VAE $\mathbb{R}^4$ & 6.9057e-02 & 6.5593e-02 & 6.4047e-02 & 6.3771e-02 \\
VAE $\mathbb{R}^{10}$ & 6.8899e-02 & 6.9802e-02 & 7.0953e-02 & 6.8871e-02 \\
\cline{1-5} 
\hline
\rowcolor{atz_table1} 
with noise $\sigma$ & 0.01 & 0.05 & 0.1 & 0.5 \\
\cline{1-5}
VAE 2-Manifold & \textbf{5.9816e-02} & \textbf{6.9934e-02} & \textbf{9.6493e-02} & \textbf{4.0121e-01} \\
VAE $\mathbb{R}^2$ & 1.0120e-01 & 1.0932e-01 & 1.3154e-01 & 4.8837e-01 \\
VAE $\mathbb{R}^4$ & 6.3885e-02 & 7.6096e-02 & 1.0354e-01 & 4.5769e-01 \\
VAE $\mathbb{R}^{10}$ & 7.4587e-02 & 8.8233e-02 & 1.2082e-01 & 4.8182e-01 \\
\end{tabular}
\end{center}
\caption{\textbf{Manifold Latent Variable Model: VAE Reconstruction Errors} The $L^2$-relative errors of reconstruction for our VAE methods.  The final is the lowest value during training.  The manifold latent spaces show improved learning.  When an incompatible topology is used, such as $\mathbb{R}^2$, this can result in
deterioration in learned representations. With noise in the input $\tilde{X} = X + \sigma \eta(0,1)$ and reconstructing the target $X$, the manifold latent spaces also show improvements for learning.
}
\label{table:VAE_Manifold_Reconstruction}
\end{table}

To learn more parsimonous and robust representations of physical systems, we develop methods for latent spaces having geometries and topologies more general than euclidean space.  This is helpful in capturing inherent structure such as periodicities or other symmetries.  
We consider physical systems with constrained mechanics, such as the arm mechanism for reaching for objects in figure~\ref{fig:manifold_lvms}.  The observations are taken to be the two locations $\mb{x}_1, \mb{x}_2 \in \mathbb{R}^2$ giving $\mb{x} = (\mb{x}_1,\mb{x}_2) \in \mathbb{R}^4$.  When the segments are rigidly constrained these configurations lie on a manifold (torus).  We can also allow the segments to extend and consider more exotic constraints such as the two points $\mb{x}_1, \mb{x}_2$ must be on a klein bottle in $\mathbb{R}^4$.  Related situations arise in other areas of imaging and mechanics, such as in pose estimation and in studies of visual perception
~\cite{Carlsson_Klein_Bottle_Images_2014,
Carlsson2008,
Sarafianos_Review_Pose_Estimation_2016}.  For the arm mechanics, we can use this prior knowledge to construct a torus latent space represented by the product space of two circles $S^1 \times S^1$.  
To obtain a learnable class of manifold encoders, we use the family of maps 
$\mathcal{E}_\theta = \Lambda(\tilde{\mathcal{E}}_\theta(x))$, 
with $\tilde{\mathcal{E}}_\theta(x)$ into $\mathbb{R}^4$ 
and  
$\Lambda(\mb{w}) = \Lambda(w_1,w_2,w_3,w_4) = (z_1,z_2,z_3,z_4) = \mb{z}$, 
where $(z_1,z_2) = (w_1,w_2)/\|(w_1,w_2)\|$, $(z_3,z_4) = (w_3,w_4)/\|(w_3,w_4)\|$, 
see VAE Section and Appendix A.  For the case of klein bottle constraints, we use our point-cloud representation of the non-orientable manifold with the parameterized embedding in $\mathbb{R}^4$ 
\begin{eqnarray*}
\begin{array}{ll}
z_1 = (a + b\cos(u_2))\cos(u_1) & 
z_2 = (a + b\cos(u_2))\sin(u_1) \\
z_3 = b\sin(u_2)\cos\left(\frac{u_1}{2}\right) &
z_4 = b\sin(u_2)\sin\left(\frac{u_1}{2}\right),
\end{array}
\end{eqnarray*}
with $u_1,u_2 \in [0,2\pi]$.  The $\Lambda(\mb{w})$ is taken to be the map to the nearest point of the manifold $\mathcal{M}$, which we compute numerically along with the needed gradients for backpropogation as discussed in Appendix A.  

Our VAE methods are trained with encoder and decoder DNN's having layers of sizes (in)-100-500-100-(out)
with Leaky-ReLU activations with s = 1e-6 with results reported in Figure~\ref{fig:manifold_lvms} and Table~\ref{table:VAE_Manifold_Reconstruction}.  We find
learning representations is improved by use of the manifold latent spaces, in these trials even showing a slight edge over $\mathbb{R}^4$.  When the wrong topology is used, such as in $\mathbb{R}^2$, we find in both cases a significant deterioration in the reconstruction accuracy, see Table~\ref{table:VAE_Manifold_Reconstruction}.  This arises since the encoder must be continuous and hedge against the noise regularizations.  This results in an incurred penalty for a subset of configurations.  The encoder exhibits non-injectivity and a rapid doubling back over the space to accommodate the decoder by lining up nearby configurations in the topology of the input space manifold to handle noise perturbations in $z$ from the probabilistic nature of the encoding.  We also studied robustness when training with noise for $\tilde{X} = X + \sigma\eta(0,1)$ and measuring accuracy for reconstruction relative to target $X$.  As the noise increases, we see that the manifold latent spaces improve reconstruction accuracy acting as a filter through restricting the representation.  The probabilistic decoder will tend to learn to estimate the mean over samples of a common underlying configuration and with the manifold latent space restrictions is more likely to use a common latent representation.  For $\mathbb{R}^d$ with $d > 2$, the extraneous dimensions in the latent space can result in overfitting of the encoder to the noise.  We see as $d$ becomes larger the reconstruction accuracy decreases, see Table~\ref{table:VAE_Manifold_Reconstruction}.  These results demonstrate how geometric priors can aid learning in constrained mechanical systems.

\section{Conclusions}
We developed VAE's for learning robustly nonlinear dynamics of physical systems by introducing methods for latent representations utilizing general geometric and topological structures.  We demonstrated our methods for learning the non-linear dynamics of PDEs and constrained mechanical systems.  We expect our methods can also be used in other physics-related tasks and problems to leverage prior geometric and topological knowledge for improving learning for nonlinear systems.

\section{Acknowledgments}
{ \fontsize{9.0}{10} \selectfont
Authors research supported by grants DOE Grant ASCR PHILMS DE-SC0019246 and NSF Grant DMS-1616353. Also to R.N.L. support by a donor to UCSB CCS SURF program.  Authors also acknowledge UCSB Center for Scientific Computing NSF MRSEC (DMR1121053) and UCSB MRL NSF CNS-1725797.  P.J.A. would also like to acknowledge a hardware grant from Nvidia. 
}

\newpage
\clearpage

\bibliography{paper_database.bib}

\newpage
\clearpage

\section{Appendix A: Backpropogation of Encoders for Non-Euclidean Latent Spaces given by General Manifolds}

\begin{figure}[ht]
\centerline{\includegraphics[width=0.99\columnwidth]{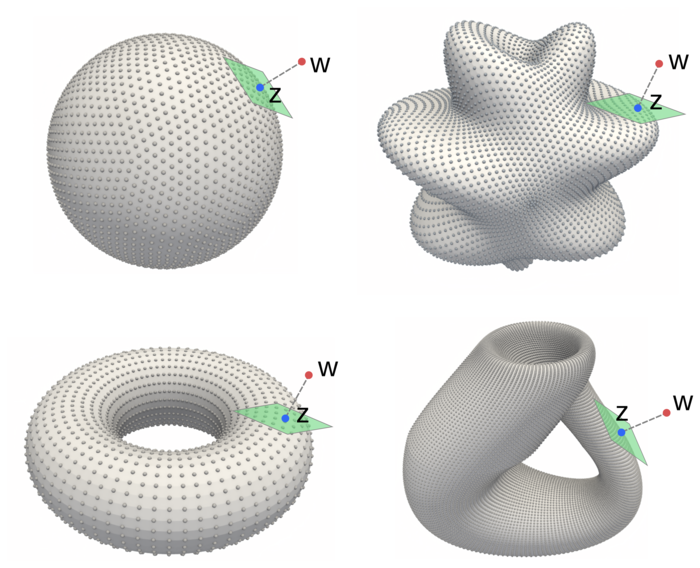}}
\caption{\textbf{Learnable Mappings to Manifold Surfaces}  We develop methods based on point cloud representations embedded in $\mathbb{R}^{n}$ for learning latent manifold representations having general geometries and topologies. 
}
\label{fig:manifold_map}
\end{figure}

We develop methods for using backpropogation to learn encoder maps from $\mathbb{R}^d$ to general manifolds 
$\mathcal{M}$.  We perform learning using the family of manifold encoder maps of the form
$\mathcal{E}_\theta = \Lambda(\tilde{\mathcal{E}}_\theta(x))$.  This allows for use of latent spaces having general topologies and geometries.  We represent the manifold as an embedding $\mathcal{M} \subset \mathbb{R}^{2m}$ and computationally use point-cloud representations along with local gradient information, see Figure~\ref{fig:manifold_map}.  To allow for $\mathcal{E}_\theta$ to be learnable, we develop approaches for incorporating our maps into general backpropogation frameworks.

For a manifold $\mathcal{M}$ of dimension $m$, we can represent it by an embedding within $\mathbb{R}^{2m}$, as supported by the Whitney Embedding Theorem~\cite{Whitney1944}.  We let $\mb{z} = \Lambda(\mb{w})$ be a mapping $\mb{w} \in \mathbb{R}^{2m}$ to points on the manifold $\mb{z} \in \mathcal{M}$.  This allows for learning within the family of manifold encoders $w = \tilde{\mathcal{E}}_\theta(x)$ any function from $\mathbb{R}^d$ to $\mathbb{R}^{2m}$.  This facilitates use of deep neural networks and other function classes.  In practice, we shall take $\mb{z} = \Lambda(\mb{w})$ to map to the nearest location on the manifold.  We can express this as the optimization problem 
$$
z^* = \arg\min_{z \in \mathcal{M}} \frac{1}{2}\|w - z\|_2^2.
$$
We can always express a smooth manifold using local coordinate charts $\sigma^k(u)$, for example, by using a local Monge-Gauge quadratic fit to the point cloud~\cite{Atzberger_GMLS_Surf_PDE_2019}.  
We can express $z^* = \sigma^k(u^*)$ for some chart $k^*$.
In terms of the coordinate charts $\{\mathcal{U}_k\}$ and local parameterizations $\{\sigma^{k}(u)\}$ we can express this as
$$
u^*,k^* = \arg\min_{k,u \in \mathcal{U}_k} \frac{1}{2}\|w - \sigma^{k}(u)\|_2^2,
$$
where $\Phi_k(u,w) = \frac{1}{2}\|w - \sigma^k(u)\|_2^2$.  
The $w$ is the input and $u^*,k^*$ is the solution sought.
For smooth parameterizations, the optimal solution satisfies 
$$G = \nabla_z \Phi_{k^*}(u^*,w) = 0.$$  
During learning we need gradients $\nabla_w \Lambda(w) = \nabla_w z$ when $w$ is varied characterizing variations of points on the manifold $z = \Lambda(w)$.  We derive these expressions by considering variations $w = w(\gamma)$ for a scalar parameter $\gamma$.  We can obtain the needed gradients by determining the variations of $u^* = u^*(\gamma)$.  We can express these gradients using the Implicit Function Theorem as 
$$
0 = \frac{d}{d\gamma} 
G(u^*(\gamma),w(\gamma))
= \nabla_u G \frac{du^*}{d\gamma} 
+ 
\nabla_w G \frac{dw}{d\gamma}.
$$
This implies 
$$
\frac{du^*}{d\gamma} = -\left[\nabla_u G \right]^{-1}
\nabla_w G \frac{dw}{d\gamma}.
$$
As long as we can evaluate at $u$ these local gradients $\nabla_u G$, $\nabla_w G$, $dw/d\gamma$, we only need to determine computationally the solution $u^*$.  For the backpropogation framework, we use these to assemble the needed gradients for our manifold encoder maps $\mathcal{E}_\theta = \Lambda(\tilde{\mathcal{E}}_\theta(x))$ as follows.

We first find numerically the closest point in the manifold $z^* \in \mathcal{M}$ and represent it as $z^* = \sigma(u^*) = \sigma^{k^*}(u^*)$ for some chart $k^*$.  In this chart, the gradients can be expressed as
$$
G = \nabla_u \Phi(u,w) = -(w - \sigma(u))^T\nabla_u \sigma(u).
$$
We take here a column vector convention with
$\nabla_u \sigma(u) = [\sigma_{u_1} | \ldots | \sigma_{u_k}]$.  We next compute 
$$
\nabla_{u} G = \nabla_{uu} \Phi = \nabla_u\sigma^T \nabla_u \sigma - (w - \sigma(u))^T\nabla_{uu} \sigma(u)
$$
and 
$$
\nabla_w G = \nabla_{w,u} \Phi = -I \nabla_{u} \sigma(u).
$$
For implementation it is useful to express this in more detail component-wise as
$$
[G]_i = -\sum_k (w_k - \sigma_k(u))\partial_{u_i} \sigma_k(u),
$$
with 
\begin{eqnarray}
\nonumber
[\nabla_u G]_{i,j} & = & [\nabla_{uu} \Phi]_{i,j} = \sum_k \partial_{u_j} \sigma_k(u)\partial_{u_i} \sigma_k(u) \\
\nonumber
& - & \sum_k (w_k - \sigma_k(u))\partial_{u_i,u_j}^2 \sigma_k(u) \\
\nonumber
[\nabla_w G]_{i,j} & = & [\nabla_{w,u} \Phi]_{i,j} \\
\nonumber
& = & 
-\sum_k \partial_{w_j} w_k\partial_{u_i} \sigma_k(u)
= 
-\partial_{u_i} \sigma_j(u).
\end{eqnarray}
The final gradient is given by 
$$
\frac{d\Lambda(w)}{d\gamma}
= \frac{dz^*}{d\gamma} = 
\nabla_u \sigma \frac{du^*}{d\gamma} 
= -\nabla_u \sigma\left[\nabla_u G \right]^{-1}
\nabla_w G \frac{dw}{d\gamma}.
$$

In summary, once we determine the point $z^* = \Lambda(w)$ we need only evaluate the above expressions to obtain the needed gradient for learning via backpropogation
$$
\nabla_\theta \mathcal{E}_\theta (x) = 
\nabla_w \Lambda(w) \nabla_\theta \tilde{\mathcal{E}}_\theta(x), \; w = \tilde{\mathcal{E}}_\theta(x).
$$
The $\nabla_w \Lambda$ is determined by ${d\Lambda(w)}/{d\gamma}$ using $\gamma = w_1, \ldots w_n$.  In practice, the $\tilde{\mathcal{E}}_\theta(x)$ is represented by a deep neural network from $\mathbb{R}^d$ to $\mathbb{R}^{2m}$.  In this way, we can learn general encoder mappings $\mathcal{E}_\theta (x)$ from $x \in \mathbb{R}^d$ to general manifolds $\mathcal{M}$.

\end{document}